\documentclass{article}
\usepackage{spconf,amsmath,graphicx}

\usepackage{enumitem}
\usepackage{color}
\usepackage{pifont}
\setlist{nosep, leftmargin=14pt}

\usepackage{mwe} % to get dummy images

\newcommand{\cmark}{\ding{51}}%
\newcommand{\xmark}{\ding{55}}%
\title{3D Tooth Mesh Segmentation with Simplified Mesh Cell Representation}

 \name{Ananya Jana$^{\star \dagger}$, Hrebesh Molly subhash$^{\dagger}$,  Dimitris Metaxas$^{\star}$}
 \address{$^{\star}$ Department of Computer Science, Rutgers University\\  $^{\dagger}$Colgate Palmolive Company, Piscataway}
\begin{document}
%\ninept
%
\maketitle
\begin{abstract}
Manual tooth segmentation of 3D tooth meshes is tedious and there is variations among dentists.
%Manual tooth annotation of 3D tooth meshes is a tedious task.
Several deep learning based methods have been proposed to perform automatic tooth mesh segmentation. Many of the proposed tooth mesh segmentation algorithms summarize the mesh cell as - the cell center or barycenter, the normal at barycenter, the cell vertices and the normals at the cell vertices. 
Summarizing of the mesh cell/triangle in this manner imposes an implicit structural constraint and makes it difficult to  work with multiple resolutions which is done in many point cloud based deep learning algorithms. 
We propose a novel segmentation method which utilizes only the barycenter and the normal at the barycenter information of the mesh cell and yet achieves competitive performance. We are the first to  demonstrate that it is possible to relax the implicit structural constraint and yet achieve superior segmentation performance.\footnote{https://github.com/ananyajana/tooth{\_}mesh{\_}seg}

\end{abstract}

\begin{figure}[htbp]
\centering
\includegraphics[width=0.48\textwidth]{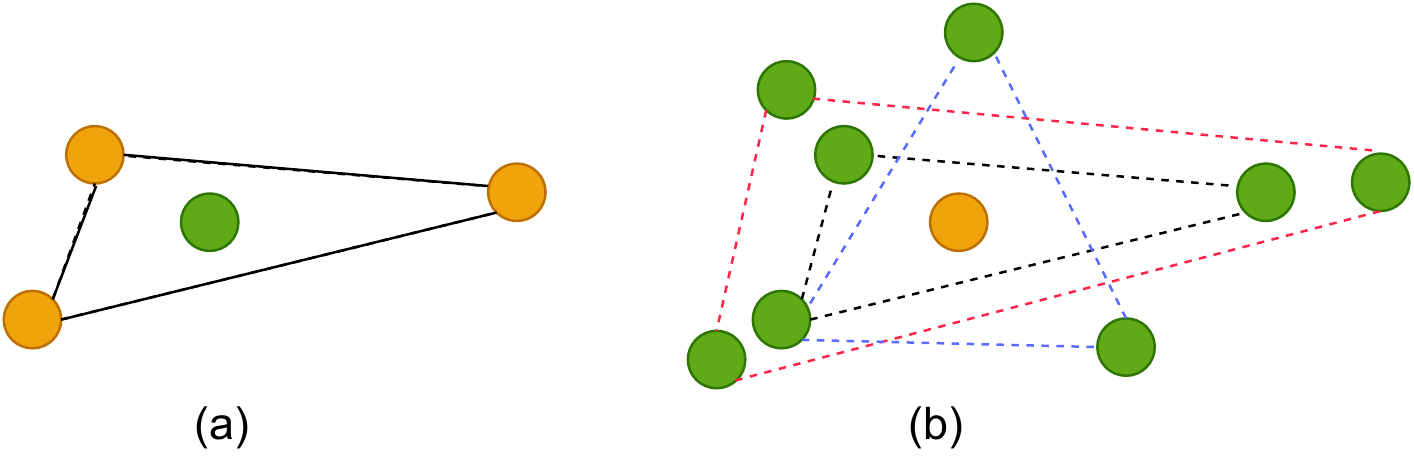}
\caption{\small The mesh cell vertices uniquely define the barycenter as shown in (a), but the barycenter does not uniquely define the mesh cell vertices as shown in (b). Utilizing the barycenter and mesh cell vertices thus impose a structural constraint. }
\label{fig:relax}
\end{figure}

\section{Introduction}
\label{sec:intro}
\begin{figure}[htbp]
\centering
\includegraphics[width=0.48\textwidth]{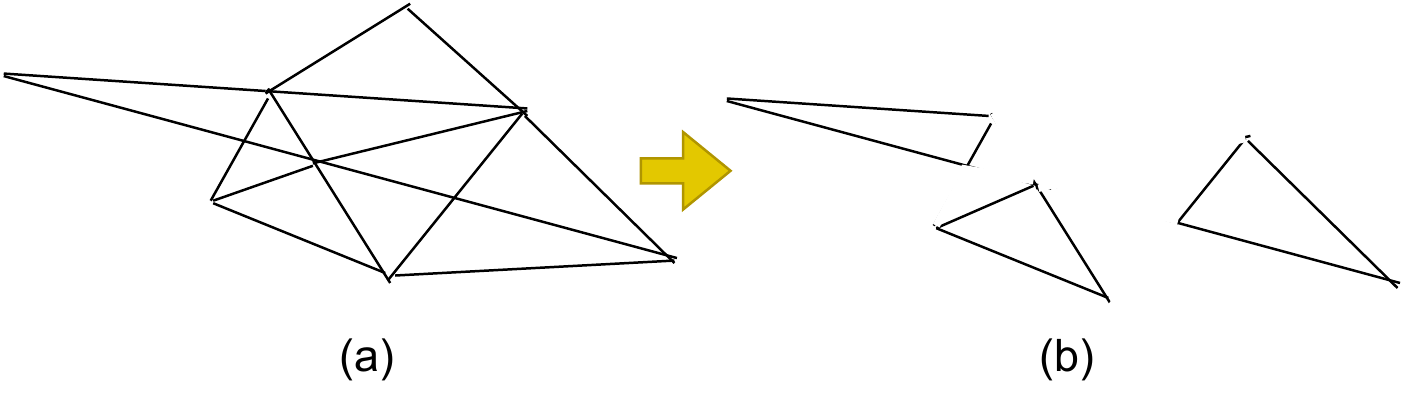}
\caption{\small (a) sample of a mesh. (b) the mesh after some triangles have been sampled randomly by sampling their barycenter. Such sampling will result in upsetting the mesh topology and loss of connectedness.}
\label{fig:downsample}
\end{figure}

With the advancement of technology, computer-aided orthodontic treatment is being widely embraced. Intraoral scanners are being widely adopted in place of the intraoral/dental cameras due to their ability to reconstruct the 3D surface. 
A vital task in computer aided orthodontic treatment is automated and accurate segmentation of teeth from intraoral scans. The intraoral scanners produce 3D surface reconstructions of the teeth either in the form of point cloud or in a mesh format or both. A highly accurate automated tooth mesh segmentation can help in downstream tasks such as recognising and classifying different dental/oral conditions like gingivitis, caries, and white lesions. There are multiple challenges involved in tooth mesh segmentation such as - crowded teeth, misaligned teeth, missing teeth. The size and shape of teeth can also vary widely across subjects. The second and third molar may evade capturing due to their being in the deep intra oral regions. Or the second/third molar might not be fully formed. Different teeth and gum conditions like recession, enamel loss etc can also alter the appearance of the teeth significantly. Multiple automatic tooth mesh segmentation algorithms have been proposed\cite{yuan2010single, zhao20213d, zhang2021tsgcnet, lian2019meshsnet, li2022multi, zhao2006interactive}. These tooth mesh segmentation algorithms can achieve  high accuracy. Some of these methods can even achieve high accuracy when trained on a single 3D tooth mesh\cite{jana2022automatic}. In this paper, we note that a dominating trend in these highly accurate deep learning based tooth segmentation methods is to summarize or represent the mesh cell in a specific way which attaches the mesh cell vertices to the barycenter of the mesh cell as features. This summarizing makes it hard to use multiple resolutions of the tooth mesh in the segmentation methods. Utilizing multiple resolutions of the data is common in point cloud processing algorithms such as BAAFNet\cite{qiu2021semantic}. Sampling from the tooth mesh is also difficult with  conventional mesh cell summarizing as it leads to loss of surface information and and causes disconnectedness as shown in Fig.~\ref{fig:downsample}. It can also be noted that the existing summarizing implicitly poses a structural constraint as shown in Fig.~\ref{fig:relax}. This structural constraint on the data is artificial. The reason is that the mesh representation consists of mesh cells which are artificially created to represent the entire object surface and the mesh cells could have been alternatively laid  out as well. In other words, it is possible to have multiple mesh cell layouts for the same 3D dental surface as the mesh cells are a way to approximate the surface.
Given this constrained representation  we explore, in this paper, if we can utilize a simplified mesh cell representation by relaxing the structural constraint, yet achieve high segmentation performance. Our key contribution is - (a) proposing a novel tooth mesh segmentation method that utilizes a simplified mesh cell representation. Our model achieves competitive performance; (b) We are the first to demonstrate that the simplified mesh cell representation can be equally or even more effective if coupled with a suitable deep learning network; (c) The simplified mesh cell representation obtained by relaxing the implicit structural constraint can pave the way for utilization of multi resolution tooth mesh in the future segmentation algorithms.
\begin{figure}[t]
\centering
\includegraphics[width=0.48\textwidth]{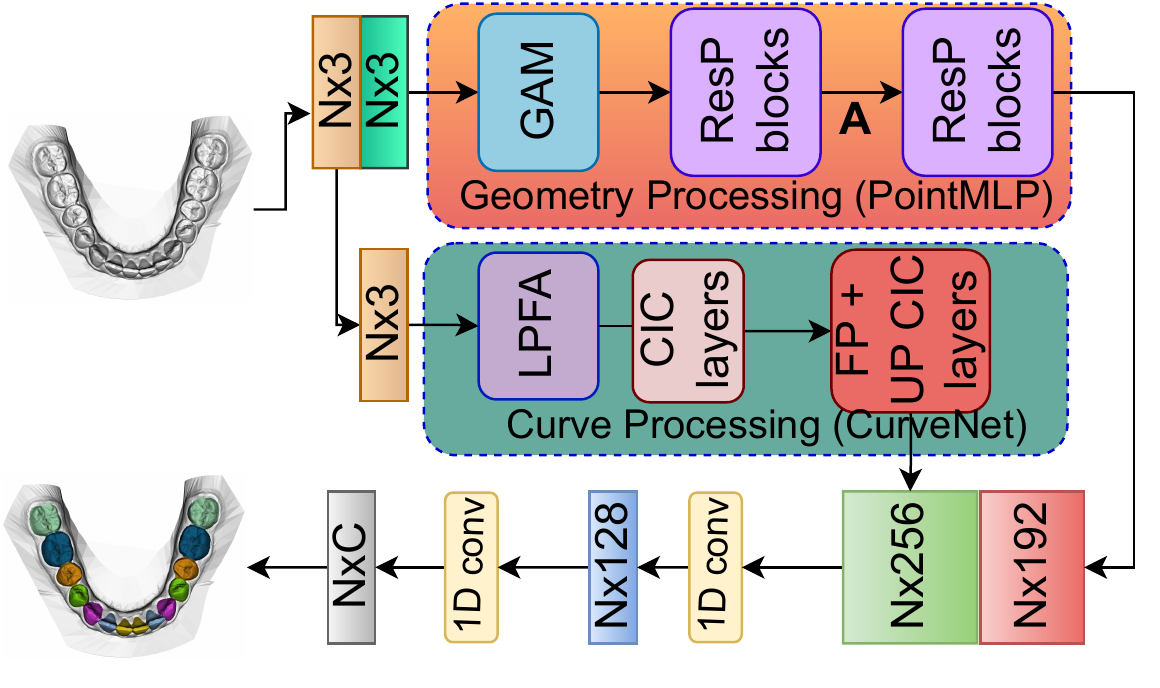}
\caption{\small Overall Architecture of our proposed network. Each mesh cell is summarized using the barycenter and the normal at the barycenter. The data is processed via a geometry processing branch and a curve processing branch}
\label{fig:network}
\end{figure}
\section{Methods}
Our proposed method has three steps (1) Data preprocessing, (2) Data augmentation, and  (3) Segmentation network to segment the jaw into the seven tooth labels and the background/gingiva label.
\subsection{Data Pre-processing}
\label{sec:preprocessing}
We utilize 589 subjects from the public dataset. These subjects do not have the wisdom teeth and hence they have a teeth count $\leq$ 14. We utilize the lower jaw scans.
Each raw lower jaw scan has labels for every point. In this work we are interested in tooth mesh segmentation hence we interpolate the pointwise labels to mesh triangle labels using k nearest neighbor algorithm. The raw lower jaw scan contains more than 100000 meshes. The meshes are downsampled to 16000 meshes using quadric downsampling. 
Each mesh cell can be characterized with four vertices - three vertices of the mesh triangle and the barycenter of the mesh triangle. With these four points, a 24 dimensional vector is constructed comprising of 12 coordinate vectors and 12 normal vectors at the four points respectively as per the convention followed in \cite{zhao20213d, zhang2021tsgcnet}. 
\subsection{Data Augmentation}
We perform three types of data augmentation to improve the model's generalization ability -  1) random rotation, 2) random translation, and
3) random rescaling. We perform 40 augmentations for each data point, thereby, effectively creating 40 new samples for each lower jaw scan.

\subsection{Segmentation Network} Our proposed method is shown in Fig.~\ref{fig:network}. Our method consists of two parallel branches - a geometry processing branch and a curve processing branch. The two branches output two different global features which are then concatenated. Finally two lightweight 1D convolutions process the concatenated global features to give the segmentation scores.\\ The current mesh cell summarizing technique utilized by the state-of-the-art methods introduces an implicit structural constraint by attaching the mesh cell vertices to the barycenter. We aim to take away this implicit constraint in our proposed method by summarizing the mesh cell with only the barycenter and the normal at the barycenter.
The relaxation in the structural constraint and the absence of the mesh vertices could potentially hamper the ability of the segmentation method in learning the representation of the mesh cell or, broadly, the representation of the surface containing the barycenter. To counter that effect we introduce the geometry processing branch in our tooth segmentation network. This geometry processing branch is a PointMLP\cite{ma2022rethinking} network and consists of a Geometric Affine Module (GAM) and a number of residual point(ResP) blocks. The geometric affine module of the PointMLP\cite{ma2022rethinking} is of interest to us as this module helps in creating a normalized representation of the surface/neighborhood even in case of sparse and diverse geometric structures. Once the vertices of the mesh cells are no longer attached to the barycenter in the form of features, the barycenters alongwith the normals at those barycenters become  sparse. The PointMLP head helps in learning representation from this comparatively sparse data and creating global feature. In addition to the geometry processing branch, we also introduce a curve processing branch in our network. We utilize CurveNet\cite{xiang2021walk} for this branch. The curve processing head is tasked with understanding and evaluating curve features from the barycenters (not on the normals) of the mesh cells. The intuition behind this step is that the different types of tooth have a large difference in shape and size, e.g. the molar teeth and the incisor teeth have different appearances. Hence, the curves induced on the barycenters coordinates (not the normals) can convey meaningful information and thereby, increase the representation learning capability of our tooth mesh segmentation network. Similar to CurveNet\cite{xiang2021walk}, the curve processing branch consists of Local Point Feature Aggregation (LPFA), Curve Intervention Convolution (CIC) follwed by feature propagation and up-Curve Intervention convolution modules.
\section{Experimental Results}
\subsection{Dataset \& Evaluation Metrics}
We use the public dataset 3D Teeth Seg Challenge 2022 \cite{ben2022teeth3ds}. The task is tooth segmentation from the 3D dental model as C = 8 different semantic parts, indicating the central incisor (T7), lateral incisor (T6), canine/cuspid (T5), 1st premolar (T4), 2nd premolar (T3), 1st molar (T2), 2nd molar (T1), and background/gingiva (BG). There are  376 subjects in the training set, 95 subjects in the validation set and 118 subjects in the test set. 
We use Dice Score(DSC), Overall Accuracy (OA), sensitivity (SEN) and Positive Predictive Value (PPV) to evaluate the performance of our model.

\begin{figure}[htbp]
\centering
\includegraphics[width=0.48\textwidth]{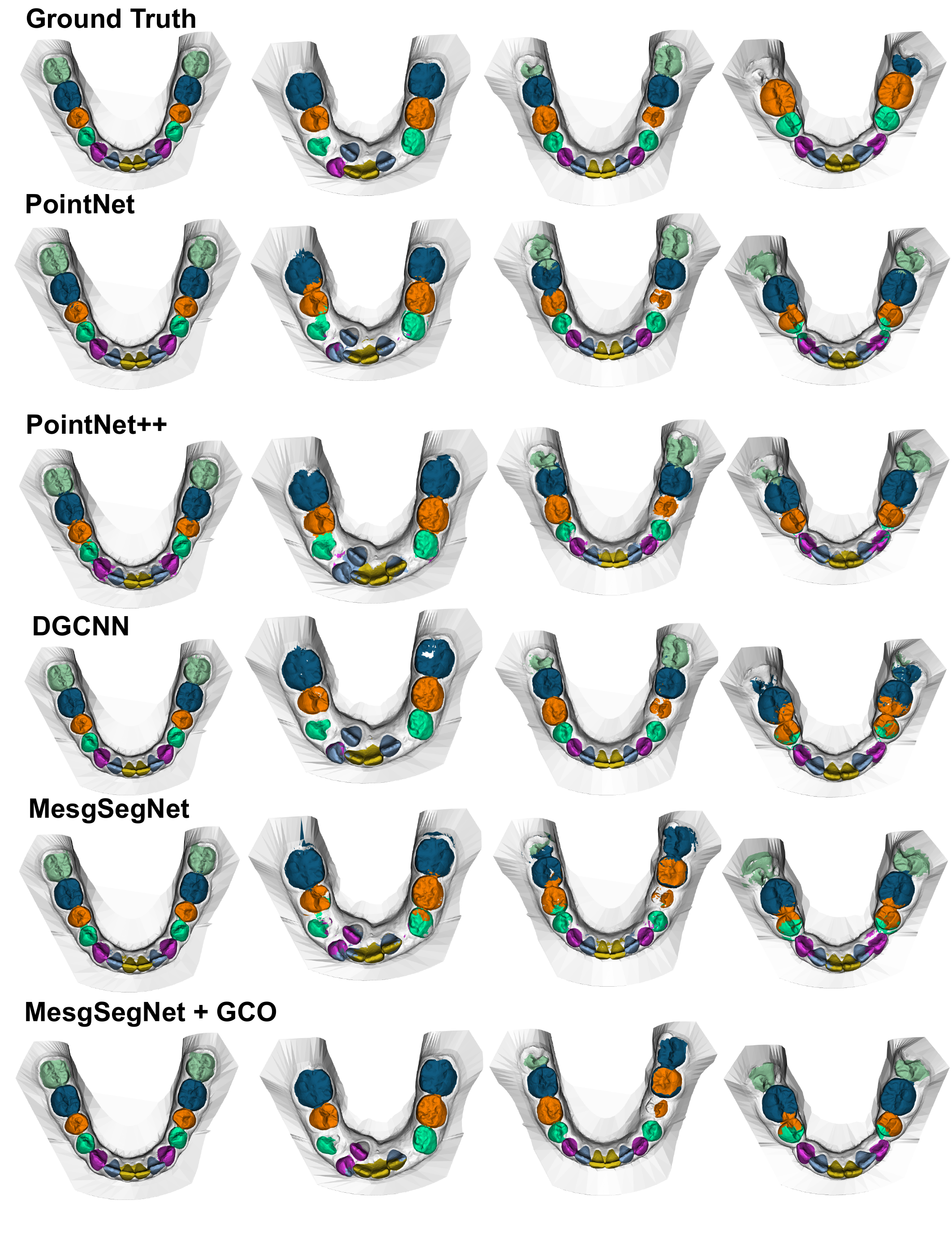}
\includegraphics[width=0.48\textwidth]{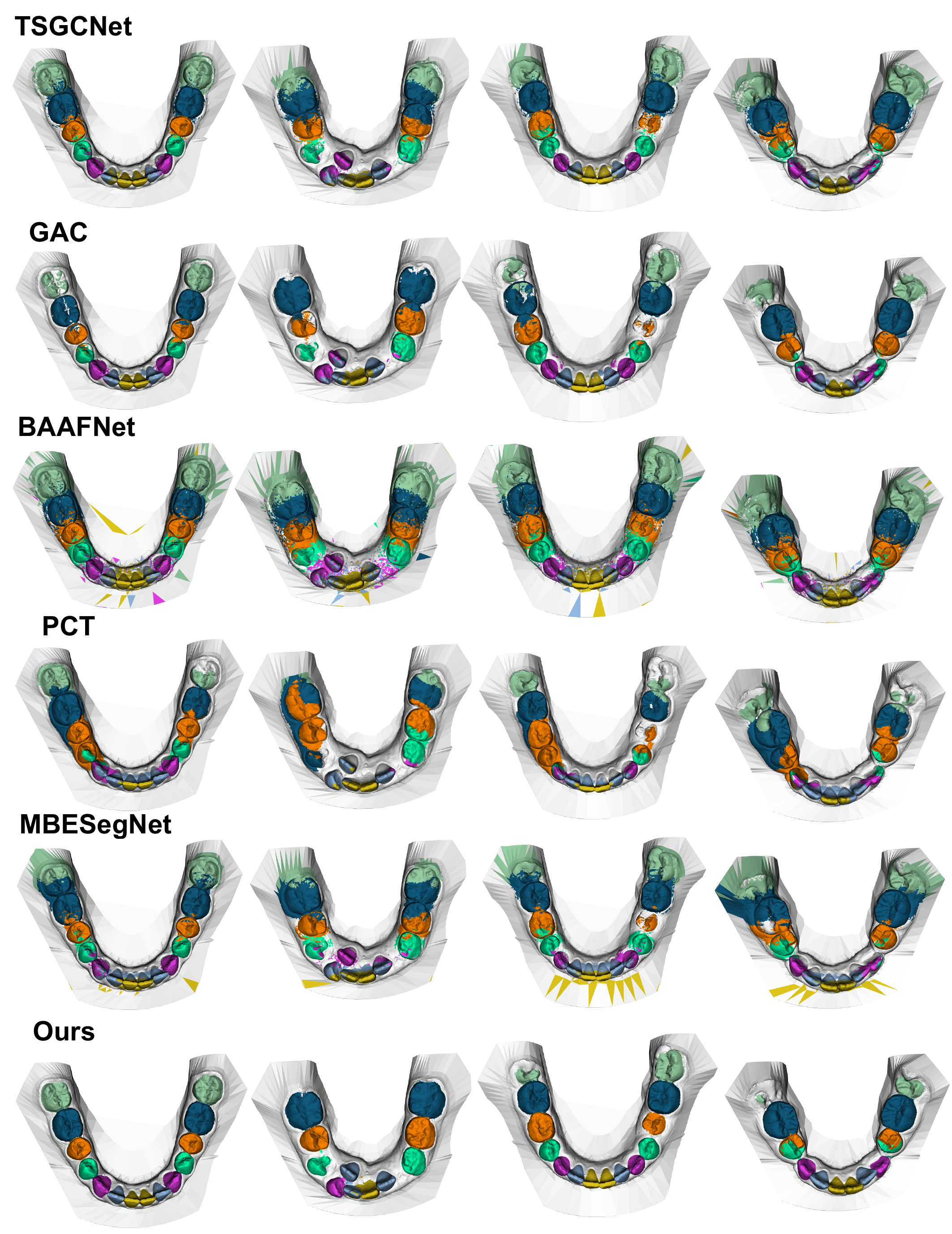}
\caption{\small The qualitative comparison of tooth labeling via different methods. Due to space constraint, we could not show all the eleven methods. (zoom in for better view in color)}
\label{fig:compareSingleTooth1}
\end{figure}

\begin{table*}[htbp]
\centering
\caption {The tooth segmentation results from ten different methods in terms of the labelwise Dice Score.}
\begin{tabular}{|c|c|c|c|c |c |c |c |c |c|}
\hline
Method   & BG & T1 & T2 & T3 &  T4 & T5 & T6 & T7  \\ \hline
PointNet [CVPR'17]\cite{qi2017pointnet} &  0.9374 & 0.7836 & 0.9100 & 0.8853 & 0.9151 & 0.8937 & 0.8994 & 0.9236\\
\hline
PointNet++ [NeurIPS'17]\cite{qi2017pointnet++}  & 0.9145 & 0.7706 & 0.8931 & 0.8663 & 0.8739 & 0.8276 & 0.7724 & 0.8275 \\ % also another one
\hline
DGCNN [ATG'19]\cite{wang2019dynamic}   & 0.9588& 0.8377& 0.9340& 0.9269& 0.9457& 0.9319& 0.9295& 0.9370\\
\hline
MeshSegNet[TMI'20]\cite{lian2020deep}    &  0.9120 & 0.7026 & 0.7899 & 0.7653 & 0.8505 & 0.8211 & 0.6744 & 0.7845\\
\hline
MeshSegNet+GCO[TMI'20]\cite{lian2020deep}  & 0.9470 & 0.8408 & 0.8948 & 0.8925 & 0.916 & 0.8690 & 0.7681 & 0.8969  \\
\hline
TSGCNet [CVPR'21]\cite{zhang2021tsgcnet}   & 0.9528& 0.6323& 0.9055& 0.9067& 0.9352& 0.9278& 0.9065& 0.9160 \\
\hline
GAC [PRL'21]\cite{zhao20213d}    & 0.8995 & 0.6330 & 0.8099 & 0.7495 & 0.8189 & 0.8365 & 0.8130 & 0.8356\\
\hline
BAAFNet [CVPR'21]\cite{qiu2021semantic}  & 0.5016 & 0.4559 & 0.6676 & 0.6293 & 0.6634 & 0.6457 & 0.5767 & 0.6724\\
\hline 
pointMLP [ICLR'22]\cite{ma2022rethinking}  &  \underline{0.9655}& \underline{0.8552}& \underline{0.9490}& \underline{0.9405}& \bf{0.9596} & \underline{0.9490}& \underline{0.9351}& \underline{0.9436}\\
\hline
PCT [CVM'21]\cite{guo2021pct}  &  0.7791 & 0.2974 & 0.5147 & 0.4496 & 0.3207 & 0.3654 & 0.4497 & 0.5788 \\
\hline
MBESegNet [ISBI'22]\cite{li2022multi}   & 0.8089 & 0.4107 & 0.6989 & 0.6852 & 0.7295 & 0.6512 & 0.5464 & 0.5255 \\
\hline
CurveNet [ICCV'21]\cite{xiang2021walk}  &  0.9540 & 0.7735 & 0.9132 & 0.9076 & 0.9291 & 0.9129 & 0.9085 & 0.9293\\
\hline
Ours & \bf{0.9657}& \bf{0.8654}& \bf{0.9516}& \bf{0.9462}& \underline{0.9595}& \bf{0.9495}& \bf{0.9395}& \bf{0.9488}\\
\hline
\end{tabular}
\label{tab:toothres}
\end{table*}

\begin{table}[t]
\centering
\caption {The tooth segmentation results from  different methods in terms of the Overall Accuracy and the Dice Score. The input column specifies how many points (p) and how many normals (n) are used in the algorithm}
\scriptsize 
\begin{tabular}{|l|c|c|c|c | c | }
\hline
Method & Input &  OA & DSC & SEN &  PPV \\ \hline
PointNet\cite{qi2017pointnet}    & 4p, 4n  &  0.9167 & 0.8935 & 0.9033 & 0.9020\\
\hline
PointNet++\cite{qi2017pointnet++}  & 4p, 4n  & 0.8820 & 0.8432 & 0.8546 & 0.8553 \\ 
\hline
DGCNN\cite{wang2019dynamic} &  4p, 4n   & 0.9435 & 0.9251 & 0.9334 & 0.9330  \\
\hline
MeshSegNet\cite{lian2020deep} & 4p, 1n   & 0.8914 & 0.8631 & 0.8787 & 0.8693 \\
\hline
MeshSegNet+GCO\cite{lian2020deep}  &  4p, 1n   &  0.9319 & 0.9085 & 0.9295 & 0.9013   \\
\hline
TSGCNet\cite{zhang2021tsgcnet} &  4p, 4n  & 0.9265 & 0.8853 & 0.9148 & 0.8928 \\
\hline
GAC\cite{zhao20213d} &   4p, 4n   &  0.8451 & 0.7994 & 0.8080 & 0.8346  \\
\hline
BAAFNet\cite{qiu2021semantic} &  4p, 4n   &  0.5910 & 0.6015 & 0.7458 & 0.5846\\
\hline
pointMLP\cite{ma2022rethinking} &  4p, 4n   & \underline{0.9537} & \underline{0.9372} & \underline{0.9468} & \underline{0.9416} \\
\hline
PCT\cite{guo2021pct} &   1p  & 0.6192 & 0.4694 & 0.4994 & 0.5760  \\
\hline
MBESegNet\cite{li2022multi} & 4p, 1n  & 0.7062 & 0.6320 & 0.7002 & 0.6344 \\
\hline
CurveNet\cite{xiang2021walk} & 1p  & 0.9298 & 0.9127 & 0.9220 & 0.9136\\
\hline
Ours & 1p, 1n &\bf{0.9553} & \bf{0.9454} & \bf{0.9505} & \bf{0.9457}\\
\hline
\end{tabular}

\label{tab:allres}
\end{table}

\begin{table}[t]
\centering
\scriptsize
\caption {The tooth segmentation results from ten different methods in terms of the Overall Accuracy and the Dice Score. b, b-n, v, v-n denote the barycenter, barycenter-normal, vertices, normals at the vertices respectively. }
\begin{tabular}{|l|c|c|c|c | c | c| c| c | c }
\hline
Method & b & b-n & v & v-n &  OA & DSC & SEN &  PPV \\ \hline
Ablation1 &   \cmark &  \cmark &  \cmark &  \cmark   & 0.9537 & 0.9372 & 0.9468 & 0.9416 \\
\hline
Ablation2 &   \cmark &  \cmark &  \xmark &  \xmark   & \underline{0.9552} & \underline{0.9405} & \underline{0.9496} & \underline{0.9435} \\
\hline
Ablation3 &   \cmark &  \xmark &  \xmark &  \xmark   & 0.9364 & 0.9157 & 0.9266 & 0.9213 \\
\hline
Ablation4 & \cmark &  \xmark &  \xmark &  \xmark   & 0.9298 & 0.9127 & 0.9220 & 0.9136\\
\hline
Ours &  \cmark &  \cmark &  \xmark &  \xmark &\bf{0.9553} & \bf{0.9454} & \bf{0.9505} & \bf{0.9457}\\
\hline
\end{tabular}

\label{tab:ablation}
\end{table}

\subsection{Implementation Details}
The model was trained using the Adam optimizer for 800 epochs with a learning rate 0.001 and  batch size 24. Cross-entropy loss was used to train the model. We select the best performing model of the 800 epochs for test. The best model was selected based on validation Dice Score (DSC).

\subsection{Results \& Discussion}
\subsubsection{Comparison with State-of-the-art}
We validate our method extensively by comparing with eleven other methods. For \cite{zhao20213d, li2022multi}, codebases were not available and hence we implemented simulations following the methods description. Out of these eleven methods, the generic point cloud segmentation methods PCT\cite{guo2021pct} and CurveNet\cite{xiang2021walk} operate on only the coordinates (1p) i.e. the barycenter of the mesh cell. MeshSegNet\cite{lian2020deep} and MBESegNet\cite{li2022multi} utilize the barycenter, normal at the barycenter and the vertices of the mesh cell (4p, 1n). The other methods utilize the 24D vector (4p, 4n) as has been described in the ~\ref{sec:preprocessing}. The results are shown in Table~\ref{tab:allres}. Our method outperforms the eleven methods. Our method is more successful compared to the other methods because of multiple factors. The 24 dimensional or even 15 dimensional mesh cell representation implicitly poses a structural constraint on the data which  is kind of artificial.  Although data augmentation tries to remedy this structural constraint, our geometry processing branch can relax this constraint more effectively with the residual connections and affine geometric module. At the same time the curve processing branch can enrich the features by adding the information regarding the curves formed using the barycenters. The curve processing branch also benefits by utilizing only the barycenter because the addition of the mesh vertices information could have confused the network. The relaxation in the structural constraint is a key advantage in our method.
\subsubsection{Ablation Study}
We performed ablation studies to illustrate the effectiveness of the proposed method. The results are shown in Table~\ref{tab:ablation}. Ablation1 is the geometry processing branch which is similar to PointMLP\cite{ma2022rethinking} but operates on the 24 dimensional vector as feature of the mesh cell. Ablation2 is similar to Ablation1 but Ablation2 only utilizes the barycenter and the normal at the barycenter. As we can see between Ablation1 and Ablation2, the relaxation of the structural constraint already has a positive effect on the geometry processing network. Ablation3 is similar to Ablation1 but Ablation3 utilizes only the barycenter and not the normal at the barycenter. This reaffirms the understanding that the normals information can encode the surface information better than just the coordinates. Ablation4 is the curve processing branch similar to CurveNet\cite{xiang2021walk}. We can see that each component of our carefully designed segmentation network improves the performance of our method.
\section{Conclusion}
In this work, we proposed a method to segment teeth from tooth mesh data using a simplified mesh cell representation. We demonstrate that although the state-of-the-art tooth segmentation methods utilize the mesh vertices as a feature of the mesh cell, this type of representation might be redundant at the commonly used resolution of the tooth mesh utilized by these state-of-the-art tooth segmentation algorithms. Rather this representation imposes an implicit structural constraint on the data which may hamper the learning and also prevent using the multi resolution of the tooth mesh data. Our proposed method based on our intuition outperforms the existing methods thus compelling us to question whether extra data always imply additional learning as generally believed, or it can be self-limiting in certain scenarios. 

\section{Compliance with Ethical Standards}
\label{sec:ethics}
This research study was conducted retrospectively using
    human subject data made available in open access by \cite{ben2022teeth3ds}. Ethical approval was not required as confirmed by the license attached with the open access data.
\section{Acknowledgments}
\label{sec:acknowledgments}
The work has been funded by the Colgate-Palmolive Company.
\bibliographystyle{IEEEbib}
\bibliography{strings,refs}

\end{document}